\documentclass[sigconf]{acmart}
\usepackage{algorithmic}
\usepackage[ruled, linesnumbered]{algorithm2e}
\newlength\mylen

\usepackage{diagbox}
\usepackage{multirow}
\AtBeginDocument{%
  \providecommand\BibTeX{{%
    \normalfont B\kern-0.5em{\scshape i\kern-0.25em b}\kern-0.8em\TeX}}}
\usepackage{array}
\usepackage{subfigure}

\copyrightyear{2023} 
\acmYear{2023} 
\setcopyright{acmlicensed}\acmConference[CIKM '23]{Proceedings of the 32nd ACM International
Conference on Information and Knowledge Management}{October 21--25, 2023}{Birmingham, United
Kingdom}
\acmBooktitle{Proceedings of the 32nd ACM International Conference on Information and Knowledge
Management (CIKM '23), October 21--25, 2023, Birmingham, United Kingdom}
\acmPrice{15.00}
\acmDOI{10.1145/3583780.3614933}
\acmISBN{979-8-4007-0124-5/23/10}

\begin{document}

\title{Independent Distribution Regularization for \\ Private Graph Embedding}


\author{Qi Hu}
\affiliation{%
  \institution{Department of CSE, HKUST}
  \city{Hong Kong SAR}
  \country{China}}
\email{qhuaf@connect.ust.hk}

\author{Yangqiu Song}
\affiliation{%
  \institution{Department of CSE, HKUST}
  \city{Hong Kong SAR}
  \country{China}}
\email{yqsong@cse.ust.hk}

\renewcommand{\shortauthors}{Hu and Song, et al.}

\begin{abstract}
Learning graph embeddings is a crucial task in graph mining tasks. An effective graph embedding model can learn low-dimensional representations from graph-structured data for data publishing benefiting various downstream applications such as node classification, link prediction, etc. However, recent studies have revealed that graph embeddings are susceptible to attribute inference attacks, which allow attackers to infer private node attributes from the learned graph embeddings. To address these concerns, privacy-preserving graph embedding methods have emerged, aiming to simultaneously consider primary learning and privacy protection through adversarial learning. However, most existing methods assume that representation models have access to all sensitive attributes in advance during the training stage, which is not always the case due to diverse privacy preferences. Furthermore, the commonly used adversarial learning technique in privacy-preserving representation learning suffers from unstable training issues. In this paper, we propose a novel approach called \textit{Private Variational Graph AutoEncoders} (PVGAE) with the aid of independent distribution penalty as a regularization term. Specifically, we split the original variational graph autoencoder (VGAE) to learn sensitive and non-sensitive latent representations using two sets of encoders. Additionally, we introduce a novel regularization to enforce the independence of the encoders. We prove the theoretical effectiveness of regularization from the perspective of mutual information. Experimental results on three real-world datasets demonstrate that PVGAE outperforms other baselines in private embedding learning regarding utility performance and privacy protection. Our code is available at \url{https://github.com/HKUST-KnowComp/PrivateGraphEncoder}
\end{abstract}




\maketitle

\section{Introduction}
Graphs are commonly used to represent network-structured data such as social networks, citation networks, knowledge graphs, etc \cite{hamilton2017inductive, kipf2016variational, wang2014knowledge}. To better utilize graph data, numerous methods for graph representation learning have been proposed \cite{grover2016node2vec, kipf2016semi, hamilton2017inductive}. The goal of graph representation learning is to learn a low-dimensional node representations from raw graph data while preserving the intrinsic properties the graph and has been proven to be effective in many downstream tasks such as node classification, link prediction, etc.

Despite the success of graph representation learning, the resulting embeddings often contain unintended sensitive information that is vulnerable to attribute inference attacks and may lead to severe privacy leakage problems. For instance, in social graphs, malicious attackers can infer sensitive personal attributes (such as gender or race) of users (graph nodes) by accessing the learned graph embeddings, even if the private information is not included in the training stage \cite{gong2018attribute, duddu2020quantifying, zhang2022inference}. Therefore, there are significant privacy leakage risks if unprotected graph embeddings are collected by malicious attackers. With the growing attention to privacy preservation, several methods for private graph representation learning \cite{hu2022learning, li2020adversarial, wang2021privacy} have been proposed to protect private information. There are two directions to learning a private graph embedding model. One direction is based on the concept of mutual information. Those frameworks set two optimization goals that maximize the mutual information between the embedding and the primary task and minimize the sensitive information \cite{wang2021privacy, jia2018attriguard} and use adversarial learning to optimize both cost functions. The other direction draws inspiration from fair representation learning, it focuses on disentangling the embedding from the private information to generate dependence-free representations \cite{hu2022learning, liu2022fair, oh2022learning}. 

\begin{figure}[tbp]
  \centering
  \includegraphics[width=0.99\linewidth]{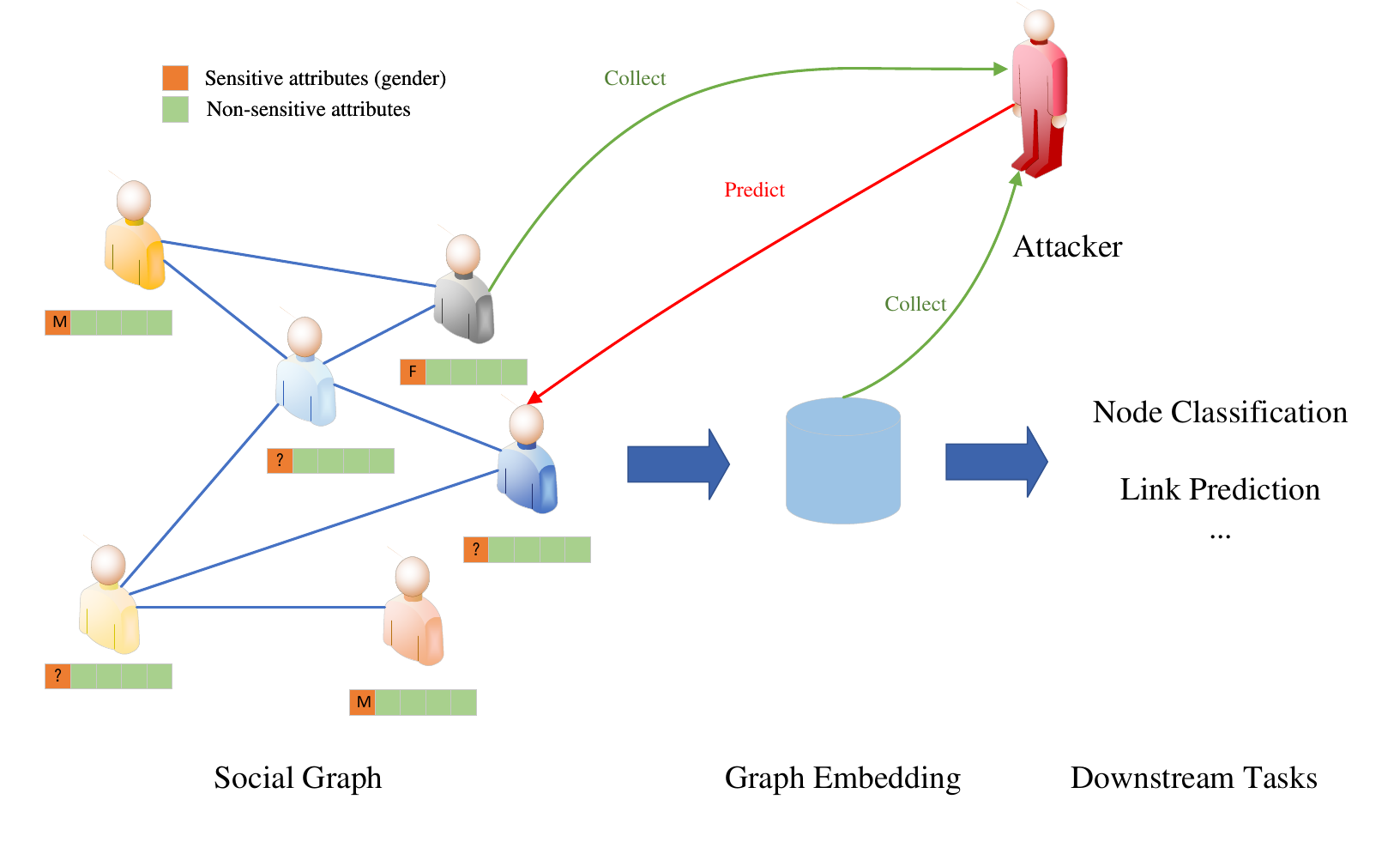}
  \caption{Overview of private graph embedding with multiple downstream tasks and different privacy preferences. Graph embeddings are generated on original graphs and applied in various downstream tasks. Users have different privacy preferences and share different attributes. Attackers infer sensitive information based on embeddings and known attributes}
  \label{fig:intro}
\end{figure}

The drawback of existing private embedding models is that they assume access to all sensitive information during the training, which is not practical in real-world applications due to varying privacy preferences. For example, as shown in Figure \ref{fig:intro}, the social network embedding can be utilized for different downstream tasks while facing the risks of attribute inference attacks. However, because users in the social network may have different privacy preferences and some users are unwilling to disclose their information to the data publisher, it is impractical to get access to all sensitive attributes for most privacy preserving methods at the training stage and learning with partially observed sensitive attributes is more common in the real world. To address this challenge, one approach proposed a semi-supervised privacy-preserving graph convolutional network \cite{hu2022learning}. However, this method solely disentangles the representations into orthogonal subspace, without considering the graph structure dependence, which influences the representative ability of the final embedding. Besides, commonly used adversarial training in privacy preservation faces unstable training problems or counter optimization objectives, which can influence the overall performance of the learned representations \cite{moyer2018invariant, liu2022fair}.  

To tackle the problems, in this paper, we propose a novel approach named private variational graph autoencoder (PVGAE) using independent distribution penalty as regularization, a framework for unsupervised learning on graph-structured data with semi-supervised privacy protection tasks. We split the original variational graph autoencoder into two parts: sensitive attribute encoder (SE) and non-sensitive encoder (NSE). SE is optimized under the instructions of sensitive attributes to estimate sensitive attribute distribution. NSE learns unsupervised graph embeddings under the independence distribution regularization to enforce distribution independence and minimize the sensitive information learned by the NSE. The NSE can be used to generate the final privacy-preserving graph representations for further various downstream tasks. In training, SE can be optimized with partially observed sensitive attributes so that PVGAE has wider application scope. Besides, the usage of distribution independence regularization avoids the adversarial training process. 

We summarize our contributions as follows:
\begin{itemize}
    \item We study the private graph representation learning problem with partially observed sensitive attributes and propose a novel privacy-preserving method PVGAE to protect the embedding from attribute inference attacks.
    \item We propose a novel independent distribution regularization for variational graph autoencoder and theoretically analyze the relation between the regularization and mutual information.
    \item Extensive experiments on three real-world datasets demonstrate that our proposed method can effectively protect the embedding from attribute inference attacks and outperforms other existing baselines.  
\end{itemize}

The rest of the paper is organized as follows. Section \ref{sec:related} systematically reviews the related work. Sectioin \ref{sec:preliminary} introduces preliminary and problem definitions. Section \ref{sec:problem_sec} first analyzes the relation between independence distribution regularization and mutual information, then introduces the framework of PVGAE in detail. Section \ref{sec:exp} evaluates the performance of PVGAE on real-world datasets. Finally, we conclude our work in Section \ref{sec:conclude}.  

\section{Related Work \label{sec:related}}
In this section, we briefly summarize the related work. Our work is closely related to graph representation learning and graph privacy preserving.
\subsection{Graph Representation Learning}
In recent years, various graph representation learning techniques have been proposed and widely applied to different real-world applications such as node classification and link prediction \cite{hamilton2017representation, zhu2020deep}. The goal of graph representation learning is to convert the raw graph to low dimensional embeddings while preserving the intrinsic graph properties \cite{chen2020graph}. Graph embeddings can reduce the space and computation overhead of downstream tasks and avoid sharing raw graph data\cite{chen2018tutorial}. There are various graph representation learning methods, including matrix factorization-based methods \cite{belkin2001laplacian, ahmed2013distributed}, random walk based algorithms \cite{grover2016node2vec, perozzi2014deepwalk}, and recently neural network based methods where graph neural networks are widely used \cite{velivckovic2017graph,hamilton2017inductive,xu2018powerful}. For example, graph convolutional network (GCN) learns graph embeddings based on spectral graph convolutions \cite{kipf2016semi}. GraphSage was proposed to learn inductive graph embeddings, graph attention neural networks introduce attention mechanism to graph embeddings \cite{hamilton2017inductive, velickovic2017graph}. Inspired by variational autoencoders (VAE) \cite{kingma2013auto}, variational graph autoencoders were proposed for unsupervised learning interpretable graph embeddings \cite{kipf2016variational, pan2018adversarially, ahn2021variational}.  

Although graph representation learning has been widely applied and has achieved great success, it faces various privacy risks. We can categorize these risks into four types \cite{hu2022learning}: membership inference attacks \cite{salem2018ml, olatunji2021membership}, model extraction attacks \cite{tramer2016stealing}, link stealing attacks \cite{he2021stealing, zhang2021link}, and attribute inference attacks \cite{gong2018attribute,duddu2020quantifying}. In membership inference attacks, the adversary aims to identify whether a node is used for model training. Model extraction attacks and link stealing attacks try to steal information of the graph representation model and original graph link, respectively. In this work, we focus on protecting graph embeddings from attribute inference attacks that exploit sensitive information of nodes.
 
\subsection{Graph Privacy Preserving}
There are several ways to preserve privacy in graph embeddings. For example, in distributed learning, federated learning can be applied to graph embedding learning, which prevents the transmission of all participants' raw data \cite{peng2021differentially, he2021fedgraphnn}. However, federated learning is not suitable for learned graph embeddings to avoid attribute inference attacks. To tackle the problems, several approaches have been proposed to generate private graph embeddings. One direction is to use adversarial training to remove sensitive information from graph embeddings \cite{wang2021privacy, liao2020graph,li2020adversarial}. These methods split the private graph embedding learning tasks into primary learning and privacy protection two sub-tasks, and perform an adversarial min-max game to minimize the sensitive information learned by the graph embeddings while preserving maximum utility information. Another direction is derived from fairness learning, these methods disentangle the graph embedding from sensitive attributes \cite{liu2022fair, oh2022learning}. Though their original goal is to learn unbiased embeddings to sensitive attributes, the motivation is also promising in private graph embeddings learning. Additionally, differential privacy can be used in graphs to protect privacy \cite{kasiviswanathan2013analyzing, day2016publishing, shen2013mining}. It provides privacy guarantees for the individual privacy of the datasets \cite{daigavane2021node}. However, the common approach in differential privacy is introducing noise into the graph embedding model, which significantly influences utility performance. 

While there are various private graph embeddings, they have the assumption that all sensitive attributes are known in the training stage, which is not always guaranteed in real-world applications. Due to the different privacy preferences of users, there may be partially sensitive attributes that can be observed limiting most methods' usage range. To address this limitation, Hu et al. \cite{hu2022learning} proposed DP-GCN to learn privacy-preserving graph representations based on GCN with partially observed sensitive attributes. However, this approach only disentangles the representations to orthogonal subspace without considering the sensitive attribute and graph structure dependence. In contrast, we propose to learn two sets of independent graph encoders to represent sensitive and non-sensitive information, respectively, with independent distribution regularization. The graph encoders utilize both attributes and graph structure information to generate high-quality private graph embeddings.  

\section{Preliminary and Problem Formulation \label{sec:preliminary}}

\subsection{Preliminary \label{subsec:pre}}
We formulate the private graph representation learning problem as a variational problem. We use $\mathcal{G} = (\mathcal{V}, \mathcal{E}, \mathbf{A}, \mathbf{X})$ denotes a graph, where $\mathcal{V}$ represents the node set, $\mathcal{V} = \{v_1, \cdots, v_N\}$, $\mathcal{E} \subseteq \mathcal{V} \times \mathcal{V} $ indicates the all edge set, $A \in \{0,1\}^{N \times N}$ is an adjacency matrix describing the graph structure where $A_{uv} = 1$ denotes there is an edge $e_{uv}$ between node $u$ and node $v$, and $A_{uv} = 0$ otherwise. $\mathbf{X} \in \mathbb{R}^{N \times D}$ denotes the feature matrix with each node has $D$ dimension features. We indicate each node $u$ with a tuple $(\mathbf{x}_u, \mathbf{s}_u, y_u)$ where $\mathbf{x}_u \in \mathbb{R}^{D-k}$ represents the non-sensitive feature vector, $\mathbf{s}_u \in \mathbb{R}^k$ is a vector of sensitive features, and $y_u \in \mathcal{Y}$ is a ground-truth label. The purpose of graph representation learning is to learn a node embedding function $f_\theta$ mapping node feature vectors in the observational space to node representations $\mathbf{Z} = \{\mathbf{z}_1, \cdots, \mathbf{z}_N\}$ by capturing the graph structural information: $\mathbf{Z} = f_\theta (\mathbf{X}, \mathbf{A}) \in \mathbb{R}^{N\times d}$, where $d$ is graph embedding dimension. 

The learned node representations can be used in various downstream tasks, i.e., node classification and link prediction. In node classification, given a set of $\mathcal{V}_L \subset \mathcal{V}$ labeled nodes with node representations $\{(\mathbf{z_u}, y_u)\}_{u \in \mathcal{V}_L}$ as training nodes, we aim to learn a node classifier $g_\phi: \mathbb{R}^d \rightarrow \mathbb{R}^{|\mathcal{Y}|}$, parameterized by $\phi$, to predict the unlabeled nodes' labels. In link prediction, given a set of positive links $\mathcal{E}_p \subset \mathcal{E}$ and a set of negative links $\mathcal{E}_n = \{e | e \notin \mathcal{E}\}$ as the training links and associated nodes' representations as input and learn a link predictor $h_\psi: \mathbb{R}^d \times \mathbb{R}^d \rightarrow [0,1]$, parameterized by $\psi$, to predict the given link whether exists in the graph.   

Variational graph autoencoder (VGAE) is an unsupervised graph representation learning method that can embed a graph into interpretable latent representations. VGAE can be decomposed into an encoder and a decoder. The encoder follows a GNN-parameterized variational posterior distribution $q_\theta (\mathbf{Z}|\mathbf{X}, \mathbf{A})$. The decoder follows a generative distribution $p_\varphi (\mathbf{A}|\mathbf{Z})$, where $\theta$ and $\varphi$ are the model parameters. Usually, The prior distribution $p(\mathbf{Z})$ assumption is used as a regularization for $q_\varphi(\mathbf{Z}|\mathbf{X}, \mathbf{A})$. VGAE is optimized by variational lower bound $\mathcal{L}$ \cite{kipf2016variational}:
\begin{equation}
    \mathcal{L} = -\mathrm{KL}[q_\theta(\mathbf{Z}|\mathbf{X}, \mathbf{A}) \| p(\mathbf{Z})]+\mathbb{E}_{q_\theta(\mathbf{Z}|\mathbf{X}, \mathbf{A})}[\log p_\varphi(\mathbf{A}|\mathbf{Z})],
\end{equation}
where $\mathrm{KL}[q(\cdot)\|p(\cdot)]$ is the Kullback-Leibler (KL) divergence between $q(\cdot)$ and $p(\cdot)$ and the second reconstruction loss evaluates the similarity between the generated graph and the input structure. VGAE is widely used in graph representation learning but faces privacy leakage problems. We aim to preserve privacy in VGAE through variational independence. 

\subsection{Problem Formulation \label{subsec:problem}}
Suppose there is a graph $\mathcal{G} = (\mathcal{V}, \mathcal{E}, \mathbf{A}, \mathcal{X})$ where part of node attributes are sensitive. The learned graph embeddings contain lots of sensitive information where an attacker can easily infer private attributes. We aim to learn graph representations for downstream tasks and protect these sensitive attributes from attribute inference attacks. However, due to different privacy preferences, we do not have access to all sensitive attributes during the representation learning stage. Following the setting from \cite{hu2022learning}, we assume that we can access part of users' sensitive attributes $S_k \subset S$ and learn node latent representations which eliminate sensitive information for various downstream tasks.

\section{Private Variational Graph AutoEncoder\label{sec:problem_sec}}

In this section, we present the private variational graph autoencoders for generating private graph embeddings. PVGAE can protect graph embeddings from attribute influence attacks, which adopts two encoders 
 $q_{\theta_x}$ and $q_{\theta_s}$ to encode nodes to independent variational representation spaces $\mathbf{Z}_x$ and $\mathbf{Z}_s$ separately with different semantic meanings. $\mathbf{Z}_s$ encodes all the private information and  guides the sensitive information removal from $\mathbf{Z}_x$, and $\mathbf{Z}_x$ maximally retains the non-sensitive information for various downstream tasks. Besides, PVGAE has two decoders $p_{\varphi_x}$ and $p_{\varphi_s}$ to decode the non-sensitive graph information and sensitive information, respectively. 
\begin{figure*}[ht]
  \centering
  \includegraphics[width=0.85\textwidth]{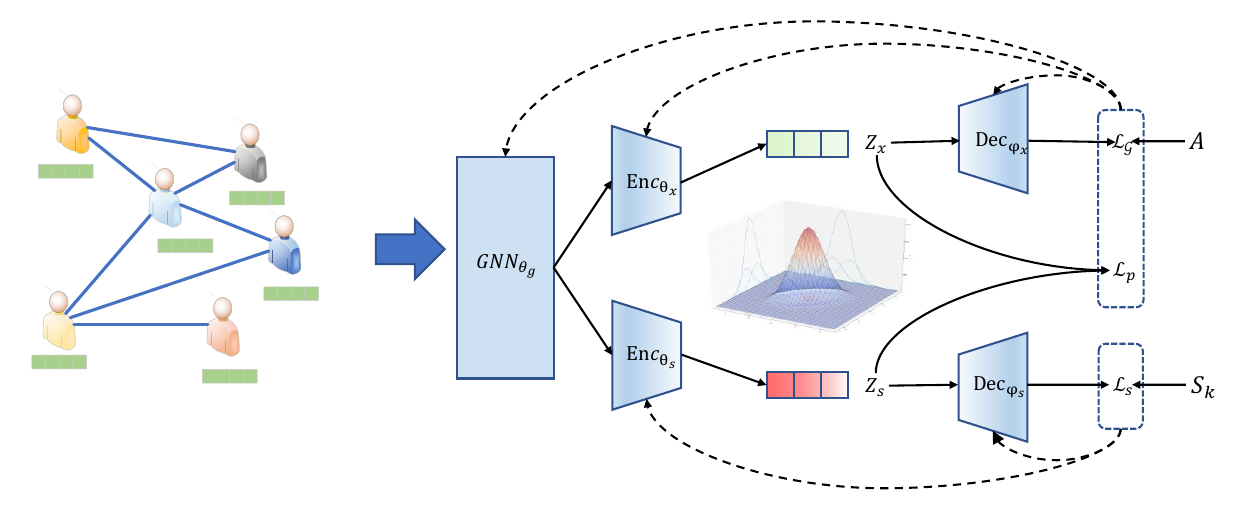}
  \caption{The overall architecture of PVGAE. Solid lines and dashed lines denote the computation process and backpropagation process respectively. The framework is a combination of two VGAEs (blue color) which learns graph structural representation $\mathbf{Z}_x$ and sensitive attribute representation $\mathbf{Z}_s$ respectively, where $\mathbf{Z}_x$ can be applied in various downstream tasks with privacy preserved. To remove the sensitive information learned by the $\mathbf{Z}_x$, we keep the two distributions independent from each other by the variational distribution penalty $\mathcal{L}_p$. Two encoders encode the same feature space generated by GNN. Because we can learn sensitive VGAE with part of sensitive attributes $\mathbf{S}_k$, our framework supports learning private representation with partially observed sensitive attributes.}
  \label{Fig:framework}
\end{figure*}

\subsection{VGAE Model \label{subsec:VGAE}}
PVGAE adopts two encoders to embed graph information and private information. We apply a graph neural network $GNN_{\theta_g}$ to learn graph information and generate preliminary representations $\mathbf{H}$:
\begin{equation}
    \mathbf{H} = GNN_{\theta_g} (\mathbf{A}, \mathbf{X}).
\end{equation}
Following \cite{kingma2013auto, hu2022learning}, two variational objectives are constructed on the learned representations, respectively. 
For graph representation learning on graph structure $\mathbf{A}$:
\begin{equation}
 \label{eq:graph}
    \mathcal{L}_x = -\mathrm{KL}[q_{\theta_x}(\mathbf{Z}_x|\mathbf{H}) \| p(\mathbf{Z}_x)]+\mathbb{E}_{q_{\theta_x}(\mathbf{Z}_x|\mathbf{H})}[\log p_{\varphi_x}(\mathbf{A}|\mathbf{Z}_x)],
\end{equation} 
where $p(\mathbf{Z}_x)$ is the prior distribution of $\mathbf{Z}_x$. For sensitive information representation learning on observed sensitive attribute $\mathbf{S}_k$: 
\begin{equation}
 \label{eq:sens}
        \mathcal{L}_s = -\mathrm{KL}[q_{\theta_s}(\mathbf{Z}_s|\mathbf{H}) \| p(\mathbf{Z}_s)]+\mathbb{E}_{q_{\theta_s}(\mathbf{Z}_s|\mathbf{H})}[\log p_{\varphi_s}(\mathbf{S}_k|\mathbf{Z}_s)],
\end{equation}
where $p(\mathbf{Z_s})$ is prior distribution of $\mathbf{Z}_s$. Two objective functions consist of two components respectively. The KL divergence is the distribution distance constraint for $q_{\theta_x}(\mathbf{Z}_x|\mathbf{H})$ and $q_{\theta_s}(\mathbf{Z}_s|\mathbf{H})$ leading to short distance from prior distributions with $p(\mathbf{Z}_x)$ and $p(\mathbf{Z}_s)$. The remaining part in Eq. (\ref{eq:graph}) is the reconstruction term, which aims to encourage the learned representations to preserve the graph structure. The second term in Eq. (\ref{eq:sens}) is to estimate sensitive attribute distribution with a given graph structure and non-sensitive attributes. We Follow common assumptions for autoencoders \cite{kipf2016variational}, assuming that the prior distribution $p(\mathbf{Z}_x)$ and $p(\mathbf{Z}_s)$ to be normal distributions: 
\begin{equation}
    \begin{split}
        \mathbf{Z}_x &\sim \mathcal{N}(\mu_x, \sigma_x),\\ 
        \mathbf{Z}_s &\sim \mathcal{N}(\mu_s, \sigma_s),
    \end{split}
\end{equation}

By optimizing two objectives, we can learn two representations with various usage, $\mathbf{Z}_x$ contains graph information and has the capability for various downstream tasks such as node classification and link prediction while $\mathbf{Z}_s$ learns sensitive information and can predict private attributes. Though $\mathbf{Z}_x$ learning does not access sensitive information $S$, the potential relevance between sensitive information and graph structure, nodes' non-sensitive attributes results in privacy leakage risks. Therefore, we assume that a correlation between sensitive and non-sensitive exists and $(\mathbf{Z}_x, \mathbf{Z}_s)$ follows bivariate distributions:
\begin{equation}
    (\mathbf{Z}_x, \mathbf{Z}_s) \sim \mathcal{N}(\mu_x,\mu_s, \sigma_x, \sigma_s,\rho),
\end{equation}
where $\rho$ is the correlation coefficient between $\mathbf{Z}_x$ and $\mathbf{Z}_s$.

\subsection{Variational Independence \label{subsec:penalty}}
\begin{figure}[tbp]
  \centering
  \includegraphics[width=1\linewidth]{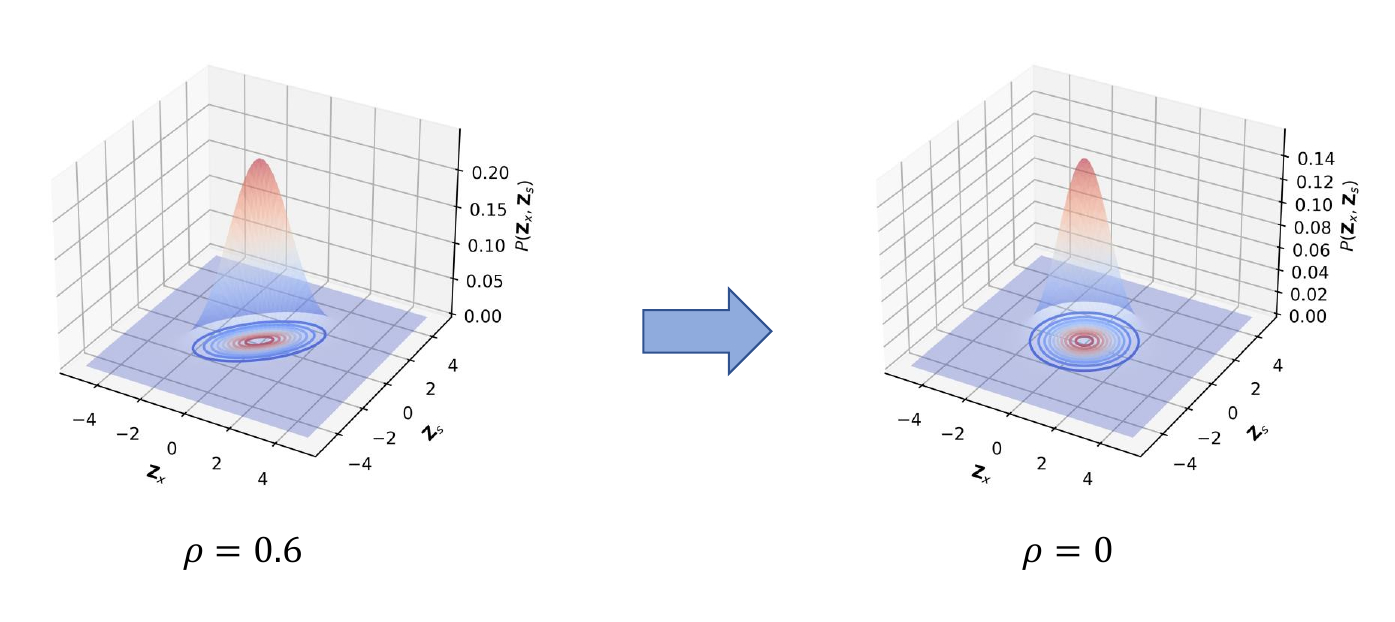}
  \caption{The illustration of distribution independence. For bivariate normal distribution, correlation indicates the intrinsic relation between two distributions. We disentangle the representations through variational distribution Independence.}
  \label{fig:distribution}
\end{figure}

 PVGAE disentangles the two learned latent distributions to eliminate the sensitive information learned by the $\mathbf{Z}_x$. From a mutual information perspective, we want to minimize the mutual information between two learned distributions:
\begin{equation}
\label{eq:obj}
    \min_{\theta_g, \theta_x, \theta_s} \mathrm{MI}[q_{\theta_x}(\mathbf{Z}_x|\mathbf{H}), q_{\theta_s}(\mathbf{Z}_s|\mathbf{H})].
\end{equation}
The mutual information in Eq. (\ref{eq:obj}) is difficult to optimize. We transform the mutual information into correlation problems. In the bivariate case, the expression for the mutual information of two normal distribution variables can be expressed as:
\begin{equation}
\begin{split}
  \label{eq:mi}
    \mathrm{MI} &= H(\mathbf{Z}_x) + H(\mathbf{Z}_s) - H(\mathbf{Z}_x, \mathbf{Z}_s) \\ 
    &= -\frac{1}{2} \ln (1 - (\frac{\sigma_{x,s}}{\sigma_x\sigma_s})^2) \\ 
    &= -\frac{1}{2} \ln (1 - \rho^2)  .
\end{split}
\end{equation}
 From Eq. (\ref{eq:mi}), we can know that mutual information is a monotonic transformation of the correlation square $\rho^2$. Therefore, the optimization objective in Eq. (\ref{eq:obj}) is equal to optimizing the correlation $\rho \to 0$. As shown in Figure \ref{fig:distribution}, the distribution for two variables is related and leads to sensitive attribute information leakage. To protect privacy, we want to minimize the mutual information between two distributions to disentangle the representations.

As we follow the assumption that the two prior distributions are bivariate distributions, we construct an auxiliary variable $\mathbf{Z}_a$ as follow:
\begin{equation}
    \mathbf{Z}_a = \mathbf{Z}_x + \mathbf{Z}_s \sim \mathcal{N}(\mu_x + \mu_s, \sqrt{\sigma_x^2 + \sigma_s^2 - 2 \rho \sigma_x \sigma_s}),
\end{equation}

To minimize the mutual information learned by the two distributions, we transform the problem to minimize the distance between the auxiliary distribution and an objective distribution:
\begin{equation}
     \min_{\theta_g, \theta_x, \theta_s}\mathrm{KL}[p(\mathbf{Z}_a), p_{\rho=0}(\mathbf{Z}_a)],
\end{equation}
the distance optimization objective is consistent with the mutual information minimization, where the lower bound are both achieved when $\rho = 0$.
Following the common assumption that the prior distribution is standard normal distribution \cite{kipf2016variational}, the mutual information objective can be transformed into KL divergence to disentangle the two distributions:
\begin{equation}
\begin{split}   
\label{eq:penalty}
 \min_{\theta_x, \theta_g} \mathcal{L}_p =  &\min_{\theta_x, \theta_g} \mathrm{MI}[q_{\theta_x}(\mathbf{Z}_x|\mathbf{H}), q_{\theta_s}(\mathbf{Z}_s|\mathbf{H})]\\
\Leftrightarrow  &\min_{\theta_x, \theta_g} \mathrm{KL}[q_{\theta_x, \theta_s}(\mathbf{Z}_a/\sqrt{2}|\mathbf{H}), \mathcal{N}(0, I)],
\end{split}
\end{equation}
where $\mathcal{L}_p$ is the distribution penalty which forces $q_{\theta_x}(\mathbf{Z}_x|\mathbf{H})$ to be independent from sensitive distribution $q_{\theta_x}(\mathbf{Z}_s|\mathbf{H})$, acting as variational independence regularization.

\subsection{Training Algorithm}
As shown in Figure \ref{Fig:framework}, the sensitive attribute information in the graph learned by the model is removed from the private graph embeddings as PVGAE has three learning objectives: learning $\mathbf{Z}_x$ for encoding non-sensitive graph information, and the learned non-sensitive latent representation can be applied in various downstream tasks; learning $\mathbf{Z}_s$ for estimating sensitive attribute distributions based on observed sensitive attributes $\mathbf{S}_k$; and through variational independence penalty, disentangle the two distributions to be independent and to eliminate the sensitive information in $\mathbf{Z}_x$. 

\subsubsection{Training objectives}
From Sections \ref{subsec:VGAE}, \ref{subsec:penalty}, we have discussed the representation learning objectives and the representation disentangling objectives. We solve the Eq. (\ref{eq:sens}) for sensitive attribute distribution estimation by maximizing $\mathcal{L}_s$:
\begin{equation}
\begin{split}
\label{eq:objsens}
    \theta_s = \arg \max_{\theta_s} -\mathrm{KL}&[q_{\theta_s}(\mathbf{Z}_s|\mathbf{H}) \| p(\mathbf{Z}_s)] \\ &+\mathbb{E}_{q_{\theta_s}(\mathbf{Z}_s|\mathbf{H})}[\log p_{\varphi_s}(\mathbf{S}_k|\mathbf{Z}_s)].
\end{split}
\end{equation}
Combine learning objective in Eq. (\ref{eq:graph}) and variational independence penalty in Eq. (\ref{eq:penalty}), we can learn the private graph embeddings which remove sensitive information by minimizing $\mathcal{L}_G$:

\begin{equation}
\begin{split}
\label{eq:objnosens}
    \theta_g, \theta_x = \arg\min_{\theta_g, \theta_s} -\mathcal{L}_g &+ \beta \mathcal{L}_p \\
    =\arg\min_{\theta_g, \theta_s} \mathrm{KL}&[q_{\theta_x}(\mathbf{Z}_x|\mathbf{H})) \| p(\mathbf{Z}_s)] \\ 
    - &\mathbb{E}_{q_{\theta_g}(\mathbf{Z}_x|\mathbf{H})}[\log p_{\varphi_g}(\mathbf{A}|\mathbf{Z}_x)]\\
    + & \beta \mathrm{KL}[q_{\theta_x, \theta_s}(\frac{\mathbf{Z}_x +\mathbf{Z}_s}{\sqrt{2}}|\mathbf{H}), \mathcal{N}(0, I)],
\end{split}
\end{equation}
where $\beta$ is the trade-off coefficient balancing the utility and privacy protection, where the larger $\beta$ denotes the stronger privacy protection.

\subsubsection{Training process}
We jointly optimize the objectives in Eqs. (\ref{eq:objsens}) and (\ref{eq:objnosens}) using an alternating optimization schema \cite{jin2020graph, hu2022learning} to update model parameters iteratively. The overall algorithms are shown in Algorithm \ref{alg:algorithm}. For each epoch, though the sensitive encoder is based on the parameters $\theta_g$ and $\theta_s$, as the objective is to extract the sensitive information learned in preliminary representation $\mathbf{H}$ and to reduce the influence on the non-sensitive representation, we only update the parameter $\theta_s$ with $\mathcal{L}_s$ several epochs. Then, we compute $\mathcal{L}_G$ and update the parameters $\theta_g$ and $\theta_x$ together to extract the structural information in the original graphs. The independence penalty will force the learned non-sensitive encoders to be independent of sensitive information. Finally, we adopt the learned PVGAE to encode private graph embeddings:
\begin{equation}
    \mathbf{Z}_x \sim q_{\theta_x}(\mathbf{Z}|GNN_{\theta_g}(\mathbf{A},\mathbf{X})),
\end{equation}
where $\mathbf{Z}_x$ can be distributed for downstream tasks with privacy protection.
\begin{algorithm}[t]

  \KwIn{Graph $\mathcal{G}=(\mathcal{V}, \mathcal{E}, \mathbf{A}, \mathcal{X})$ \newline 
        Observed sensitive attribute $\mathbf{S}_k$ \newline  
        Epochs $E$, $E_s$, Learning rates: $\eta_s$, $\eta_G$}
  \KwOut{$\theta^*_x$,$\theta^*_s$,$\theta^*_g$ }
  \BlankLine
      \For{$i=1$ \KwTo ${E}$}{
    \tcp{Update sensitive encoders using Eq. (\ref{eq:objsens})}
    \For{$j=1$ \KwTo $E_s$}{
      $\mathcal{L}_{s} = \mathrm{KL}[q_{\theta_s}(\mathbf{Z}_s|\mathbf{H}) \| p(\mathbf{Z}_s)] $
      
      $\quad\quad-\mathbb{E}_{q_{\theta_s}(\mathbf{Z}_s|\mathbf{H})}[\log p_{\varphi_s}(\mathbf{S}_k|\mathbf{Z}_s)] $
      
      $\theta_s = \theta_s - \eta_{\theta_s}\frac{\partial \mathcal{L}_{s}}{\partial \theta_s}$
    }

      \tcp{Update non-sensitive encoders using Eq. (\ref{eq:objnosens}).}
      $\mathcal{L}_{G} = -\mathcal{L}_g +\beta\mathrm{KL}[q_{\theta_x, \theta_s}(\frac{\mathbf{Z}_x +\mathbf{Z}_s}{\sqrt{2}}|\mathbf{H}), \mathcal{N}(0, I)]$
      
      $\theta_g = \theta_g - \eta_{\theta_G}\frac{\partial \mathcal{L}_{G}}{\partial \theta_g} $
      
      $\theta_x = \theta_x - \eta_{\theta_G}\frac{\partial \mathcal{L}_{G}}{\partial \theta_x} $
       }
  \KwRet{$\theta_x, \theta_g, \theta_s$}
  \caption{The Pseudo Code of PVGAE}
  \label{alg:algorithm}
\end{algorithm}

\section{Experiments \label{sec:exp}}
In this section, we conduct experiments to evaluate our proposed PVGAE. We aim to answer the following questions:
\begin{itemize}
    \item Whether PVGAE has better utility performance and privacy protection tradeoff compared to other baselines?
    \item How does PVGAE perform with different parameters?
    \item Does PVGAE provide different privacy protection to nodes with various privacy preferences?
\end{itemize}

We first introduce the experimental setup in Section \ref{subsec:setup} and  compare the utility performance and privacy preserving ability with other baselines to validate the effectiveness and conduct ablation studies to learn PVGAE's properties to answer three questions in Sections \ref{subsec:perform}, \ref{subsec:ablation}. 

\subsection{Experimental Setup \label{subsec:setup}}

\subsubsection{Dataset description} We evaluate PVGAE's performance on three real-world datasets, including two social networks, i.e. Yale and Rochester, and one ethical dataset Credit defaulter, constructed in \cite{agarwal2021towards}. The Yale and Rochester are two datasets that collected Yale University and Rochester University user relationships and user attributes in 2005. The two networks contain 8,578 nodes, 405,450 edges, and 4,563 nodes, 167,653 edges, respectively. The Credit defaulter is a dataset of 30,000 individuals with 14 spending and payment patterns which are connected based on the similarity of features. To evaluate PVGAE's performance, we select some attributes as sensitive information that we need to protect in graph representation learning. For the Yale dataset, we consider the class year as privacy and student/faculty status (short for status) as the utility; for the Rochester dataset, the gender is regarded as privacy and class year as the utility attribute. Additionally, we evaluate the embedding performance via the link prediction problem for these two datasets. While for the Credit defaulter dataset, we aim to predict whether an individual will default on the credit card payment or not and treat an individual's marital status as a sensitive attribute. Because the Credit defaulter graph is constructed by individual similarity and its links do not have realistic relationships, we do not evaluate link prediction task performance on this dataset. Besides, as some nodes lack attribute information, we only retain nodes with needed information for convenient evaluation. For link prediction tasks, we randomly sample 10\% links as the test set and sample the same number of edges for negative examples. For node classification tasks, we randomly select 20\% nodes as the test set. 

\subsubsection{Evaluation metrics}
Following previous works \cite{wang2021privacy, hu2022learning}, we evaluate PVGAE's performance from two aspects: utility and privacy. In utility tasks, we use Area under the ROC Curve (AUC) to evaluate link prediction and accuracy to evaluate node classification, and higher metrics mean stronger representative ability. In privacy tasks, we adopt private attribute estimators to predict sensitive attributes from the embeddings, and the accuracy can be used to evaluate the information leakage problem in the embeddings. The higher the prediction accuracy is, the more severe private information leakage happens. Therefore, lower prediction accuracy is desired in privacy tasks.
\subsubsection{Baselines}
To the best of our knowledge, we select several strong privacy-preserving graph representation learning methods as our baselines. The detailed baselines are listed as follows:
\begin{itemize}
    \item VGAE \cite{kipf2016variational}: This is a framework for unsupervised learning on graph-structured data based on the variational autoencoder. Reconstruction loss and Kullback-Leibler divergence are applied to GCN \cite{kipf2016semi} learning graph representations with specified graph representations. 
    \item APGE \cite{li2020adversarial}: Based on graph autoencoder (GAE), it applies an attacker in the training stage and trains the encoders to maximize the privacy prediction loss using adversarial training.  
    \item GAE-MI \cite{wang2021privacy}: From the mutual information perspective, it proposes to maximize the utility information while minimizing the private information in the learned embeddings, and finally convert the privacy-preserving problem to a min-max game. We apply the technique to graph autoencoders to generate private embeddings.
    \item VFAE \cite{liu2022fair}: VFAE proposes to use distance covariance to learn the independent representations instead of the mutual information perspective. Though VFAE is originally designed for representation fairness, it can benefit private embedding learning.
    \item DP-GCN \cite{hu2022learning}: This is a framework learning to disentangle the sensitive representations and non-sensitive representations with orthogonal constraints. The advantage of this framework is that it still works with partially observed sensitive attributes. Compared to the original paper, we use reconstruction loss in DP-GCN for a fair comparison. 
\end{itemize} 

 Note that VGAE does not have privacy-preserving techniques, we use it as a base representation performance baseline. Besides, as some baselines (APGE, GAE-MI, VFAE) need all sensitive attributes known in advance while DP-GCN and our PVGAE do not, we have two experiment settings: learning with fully observed sensitive attributes and learning with partially observed sensitive attributes. For fair comparisons, we compare all methods with the same amount of observed sensitive attributes.
\begin{table*}
\caption{Performance on different methods in terms of utility and privacy with fully observed sensitive attributes. For utility, higher accuracy indicates better representation ability while for privacy, lower accuracy denotes better privacy protection. The best results are in bold.}
\label{tab:main}
\begin{tabular}{c|cccccccc}
\hline
\multirow{3}{*}{\textbf{Methods}} & \multicolumn{3}{c|}{Yale}                                                                                                                                                                                                                    & \multicolumn{3}{c|}{Rochster}                                                                                                                                                                                                                & \multicolumn{2}{c}{Credit defulter}                                                                                                        \\ \cline{2-9} 
                                  & \multicolumn{2}{c|}{Utility}                                                                                                                              & \multicolumn{1}{c|}{Privacy}                                                     & \multicolumn{2}{c|}{Utility}                                                                                                                                  & \multicolumn{1}{c|}{Privacy}                                                 & \multicolumn{1}{c|}{Utility}                                                 & Privacy                                                     \\ \cline{2-9} 
                                  & \multicolumn{1}{c|}{\begin{tabular}[c]{@{}c@{}}link \\ (AUC)\end{tabular}} & \multicolumn{1}{c|}{\begin{tabular}[c]{@{}c@{}}status \\ (ACC)\end{tabular}} & \multicolumn{1}{c|}{\begin{tabular}[c]{@{}c@{}}Class year\\  (ACC)\end{tabular}} & \multicolumn{1}{c|}{\begin{tabular}[c]{@{}c@{}}Link \\ (AUC)\end{tabular}} & \multicolumn{1}{c|}{\begin{tabular}[c]{@{}c@{}}Class year \\ (ACC)\end{tabular}} & \multicolumn{1}{c|}{\begin{tabular}[c]{@{}c@{}}Gender\\  (ACC)\end{tabular}} & \multicolumn{1}{c|}{\begin{tabular}[c]{@{}c@{}}Credit \\ (ACC)\end{tabular}} & \begin{tabular}[c]{@{}c@{}}Marital status \\ (ACC)\end{tabular} \\ \hline
VGAE                              & 0.893                                                                      & 0.888                                                                        & 0.858                                                                            & 0.929                                                                      & 0.882                                                                            & 0.690                                                                        & 0.795                                                                        & 0.666                                                       \\ \hline
APGE                              & 0.840                                                                      & 0.835                                                                        & 0.679                                                                            & 0.892                                                                      & \textbf{0.833}                                                                   & 0.630                                                                        & \textbf{0.785}                                                               & 0.659                                                       \\
GAE-MI                            & 0.831                                                                      & \textbf{0.849}                                                               & 0.752                                                                            & 0.910                                                                      & 0.812                                                                            & 0.659                                                                        & \textbf{0.785}                                                               & 0.683                                                       \\
VFAE                              & 0.827                                                                      & 0.841                                                                        & 0.687                                                                            & 0.913                                                                      & 0.819                                                                            & 0.582                                                                        & 0.782                                                                        & \textbf{0.622}                                              \\
DP-GCN                            & 0.817                                                                      & 0.836                                                                        & 0.686                                                                            & 0.907                                                                      & 0.811                                                                            & 0.592                                                                        & 0.772                                                                        & 0.634                                                       \\ \hline
PVGAE                             & \textbf{0.841}                                                             & 0.840                                                                        & \textbf{0.665}                                                                   & \textbf{0.916}                                                             & 0.828                                                                            & \textbf{0.566}                                                               & 0.784                                                                        & 0.623                                                       \\ \hline
\end{tabular}
\end{table*}

\begin{figure*}
\centering
\subfigure[Yale]{\includegraphics[width=0.33\textwidth]{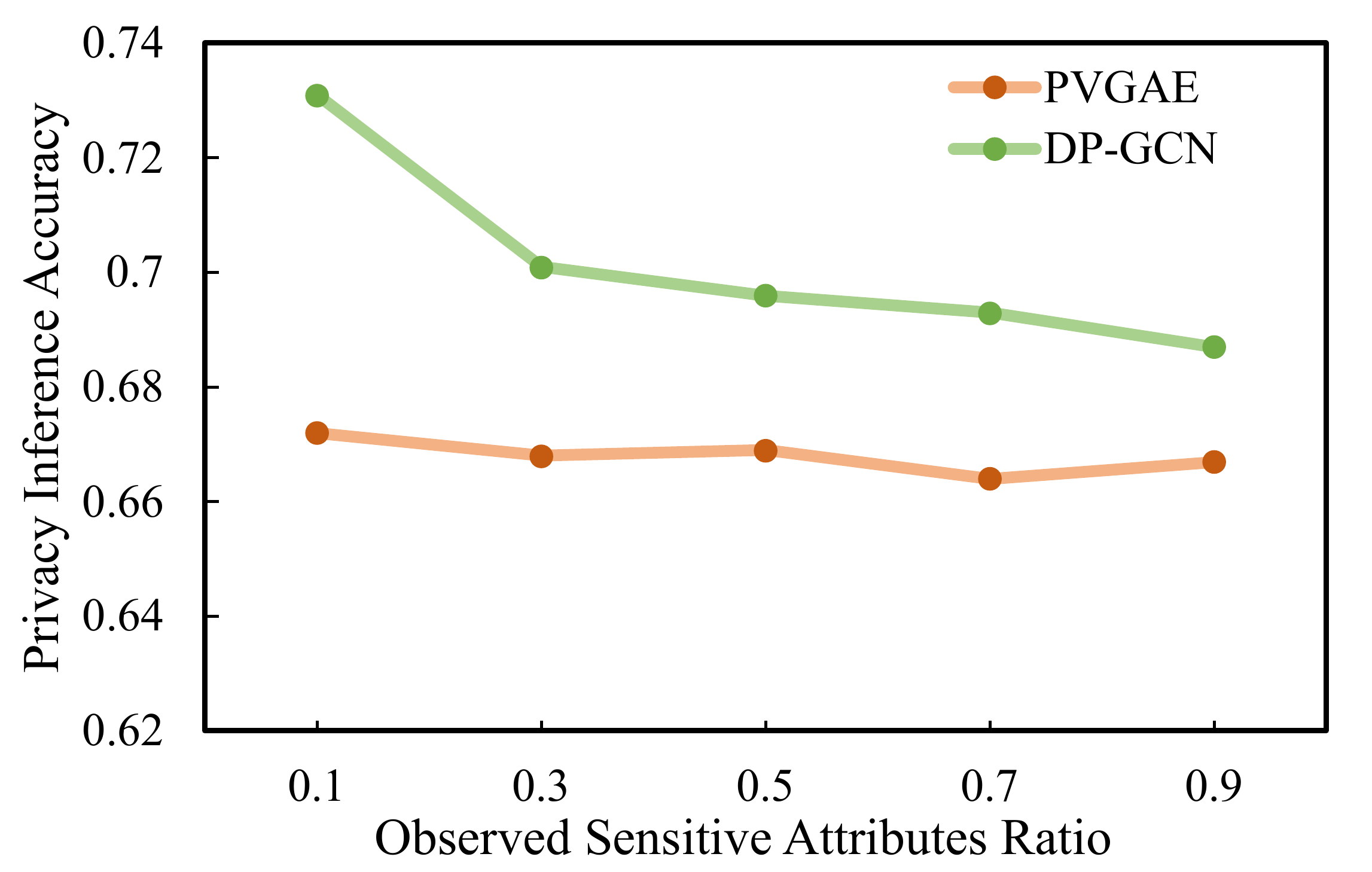}
\label{subfig:ratio-yale}}\hfil
\subfigure[Rochester]{\includegraphics[width=0.33\textwidth]{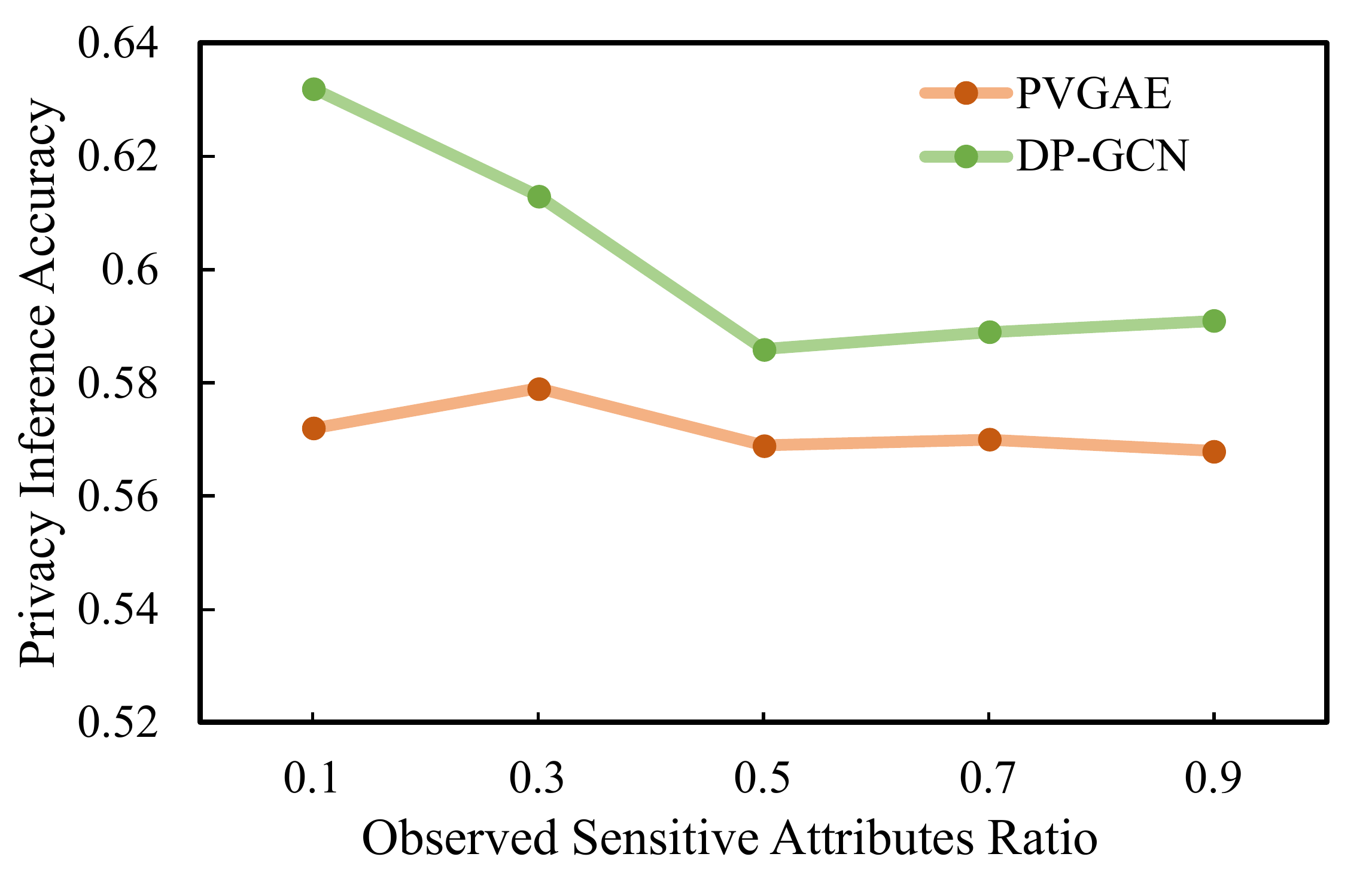}}\hfil 
\subfigure[Credit defaulter]{\includegraphics[width=0.33\textwidth]{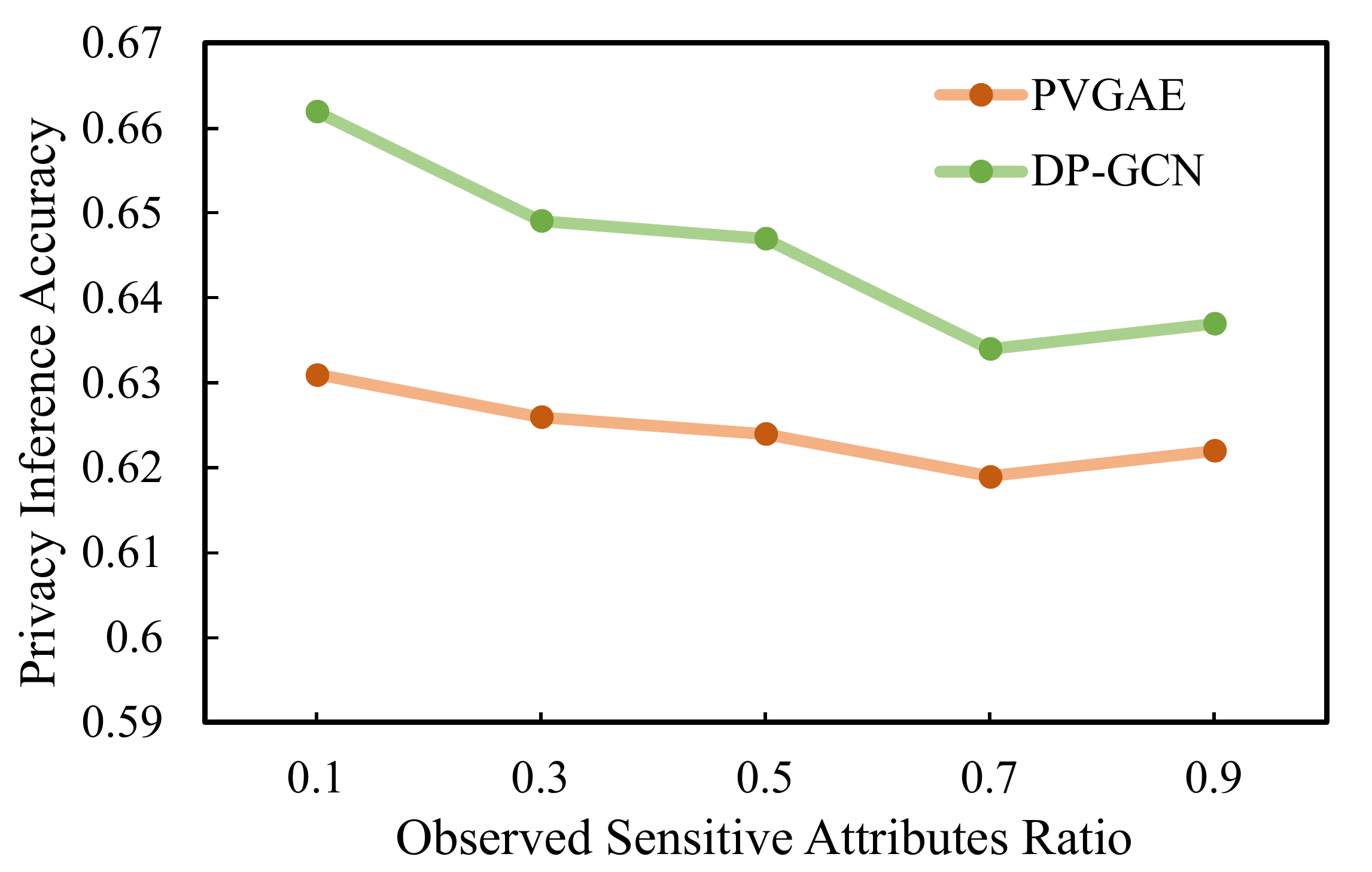}} 
\caption{Privacy preservation comparison with different ratios of observed sensitive attributes. Lower accuracy denotes better privacy protection.}
\label{fig:ratio}
\end{figure*}

\subsubsection{Parameter setting}
We train PVGAE and other baselines on the training set and tune parameters. As most baselines have privacy penalty coefficients to control trade-offs in utility performance and privacy preservation, we tune the baseline models' hyperparameters to have comparable results where models have similar utility performance and we compare the privacy inference results. For all datasets and baselines, we assume that learned embeddings are 32 dimensions, and we use Adam optimizer with an initial learning rate 0.005 for both $\eta_s$ and $\eta_G$, training epochs $E=500$ and sensitive encoder training epoch $E_s=1$. To quantify the privacy leakage problem in graph embeddings, we use MLP and cross validation from standard machine learning library \textit{scikit-learn}\footnote{\url{https://scikit-learn.org/stable/}} to predict sensitive attributes for fair evaluations.

\subsection{Performance Evaluation \label{subsec:perform}}
To evaluate the performance, we compare PVGAE with other baselines in terms of utility and privacy.

\subsubsection{Fully observed sensitive attributes} We compare PVGAE to all baselines with fully observed sensitive attributes. We assume that all private attributes are known when training the representation models. The results are shown in Table \ref{tab:main}. First, we can observe that the original VGAE without any privacy protection techniques faces severe privacy leakage problems as an attacker can easily predict sensitive information from embeddings with high accuracy. Our proposed PVGAE can effectively protect graph embeddings from attribute inference attacks. For example in Yale datasets, the attacker's prediction accuracy drops from 0.858 to 0.665 compared to the original VGAE and the finding is consistent with all datasets. Besides, though all private graph embedding methods can protect privacy effectively, they affect utility performance to varying degrees. Compared to VGAE, all privacy-aware baselines' utility drops to some extent. Our PVGAE significantly protects graph embedding privacy with a relatively slight loss of utility accuracy. Take Yale datasets as an example, PVGAE reduces the privacy inference accuracy from 0.858 to 0.665, about 23.6\%, while only decreasing the utility performance from 0.893 to 0.841 (5.9\%) and 0.888 to 0.840 (5.4\%) for link prediction and node classification, respectively. Besides, our proposed method PVGAE has competitive utility-privacy tradeoffs compared to other baselines. For example, in the Rochester dataset, PVGAE's two downstream utility tasks performances are 0.916 and 0.828, respectively, which are better than most other 
privacy-aware baselines. Meanwhile, PVGAE provides the best privacy protection as private attributes attacker accuracy is 0.566, which is lower than other baselines.  
\begin{figure*}
\centering
\subfigure[Yale]{\includegraphics[width=0.325\textwidth]{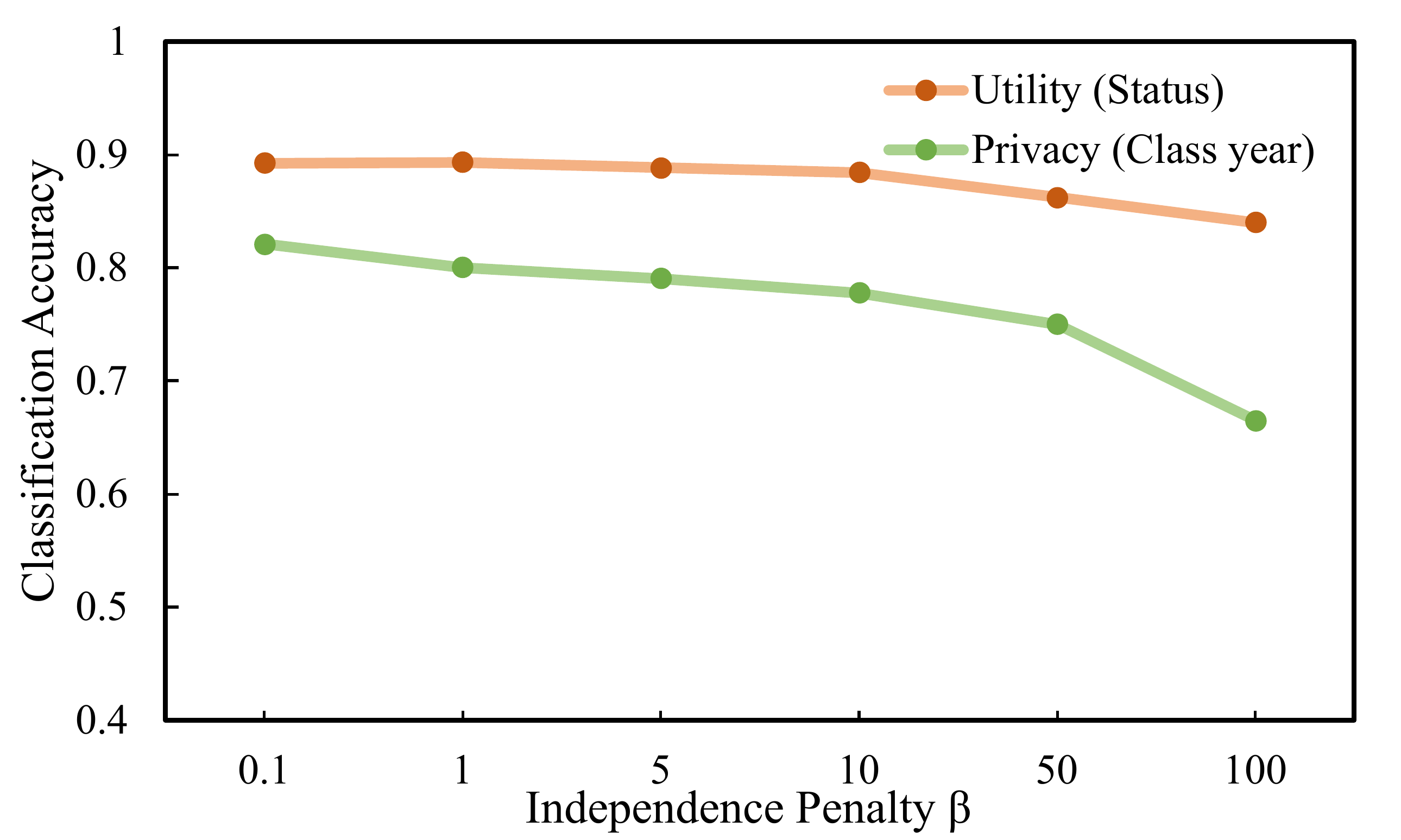}
\label{subfig:penalty-yale}}\hfil
\subfigure[Rochester]{\includegraphics[width=0.325\textwidth]{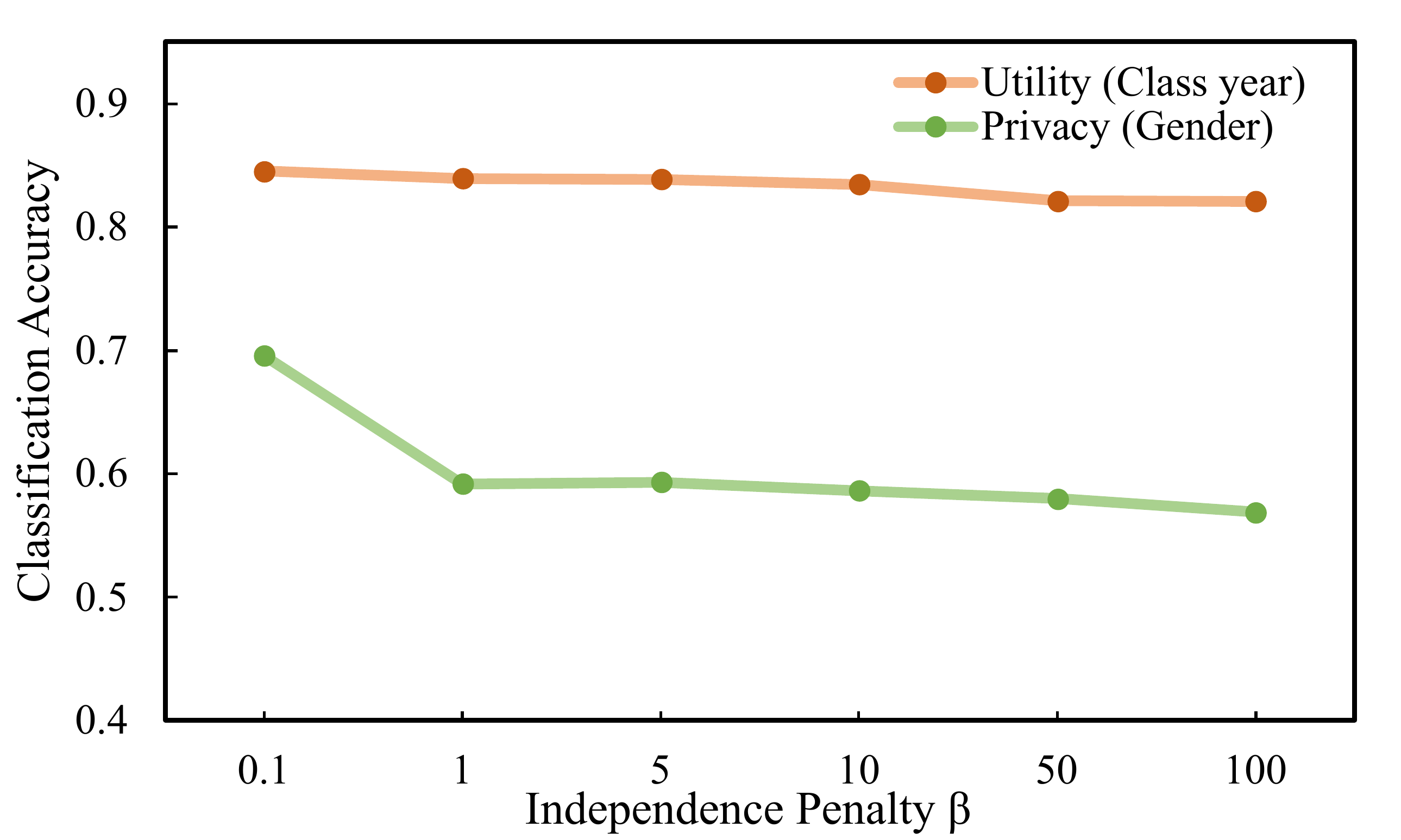} \label{subfig:penalty-rochester}}
\hfil 
\subfigure[Credit defaulter]{\includegraphics[width=0.325\textwidth]{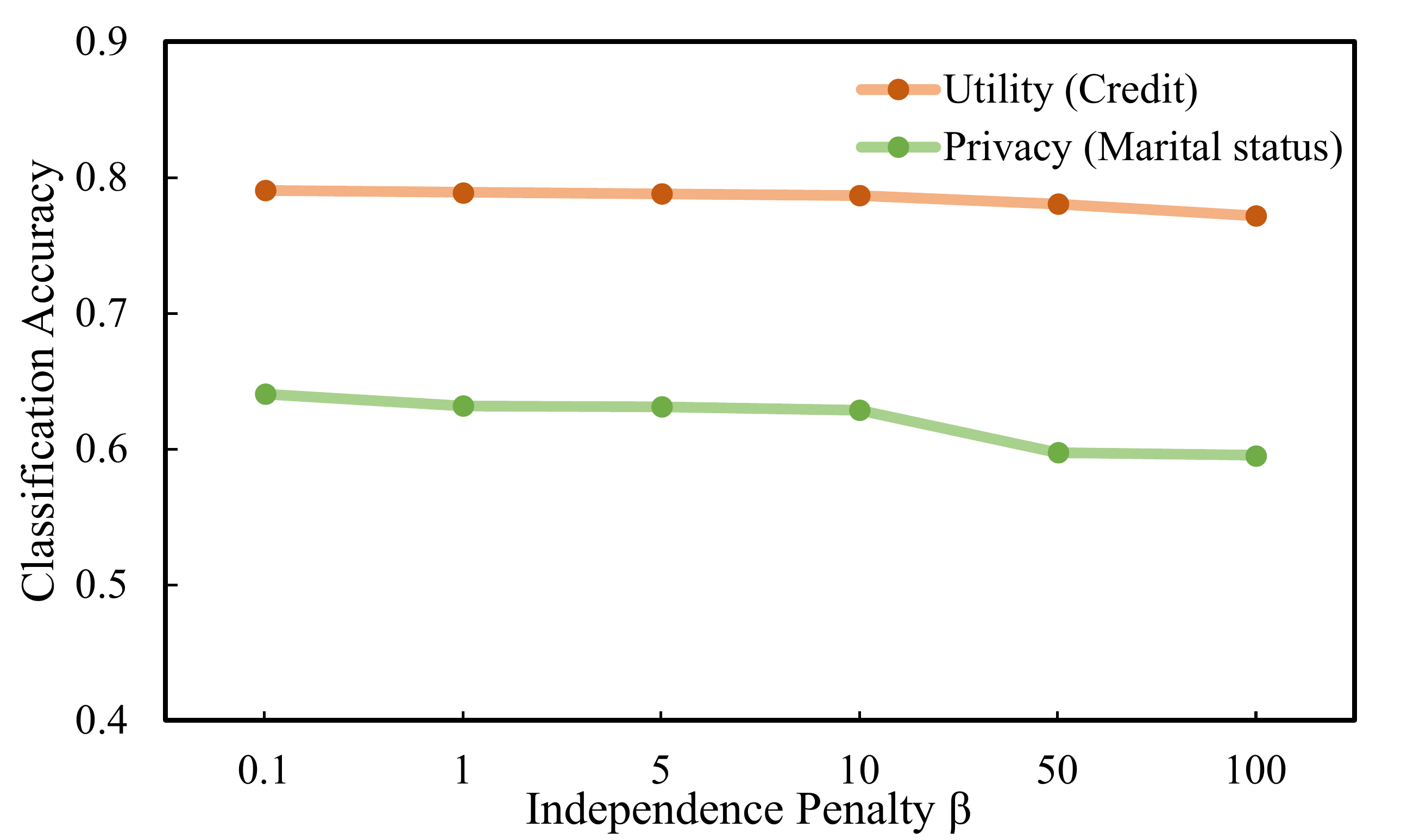}} 
\caption{Utility and privacy tradeoffs with different independence penalty coefficients.}
\label{fig:penalty}
\end{figure*}

\begin{figure*}
\centering
\subfigure[Yale]{\includegraphics[width=0.33\textwidth]{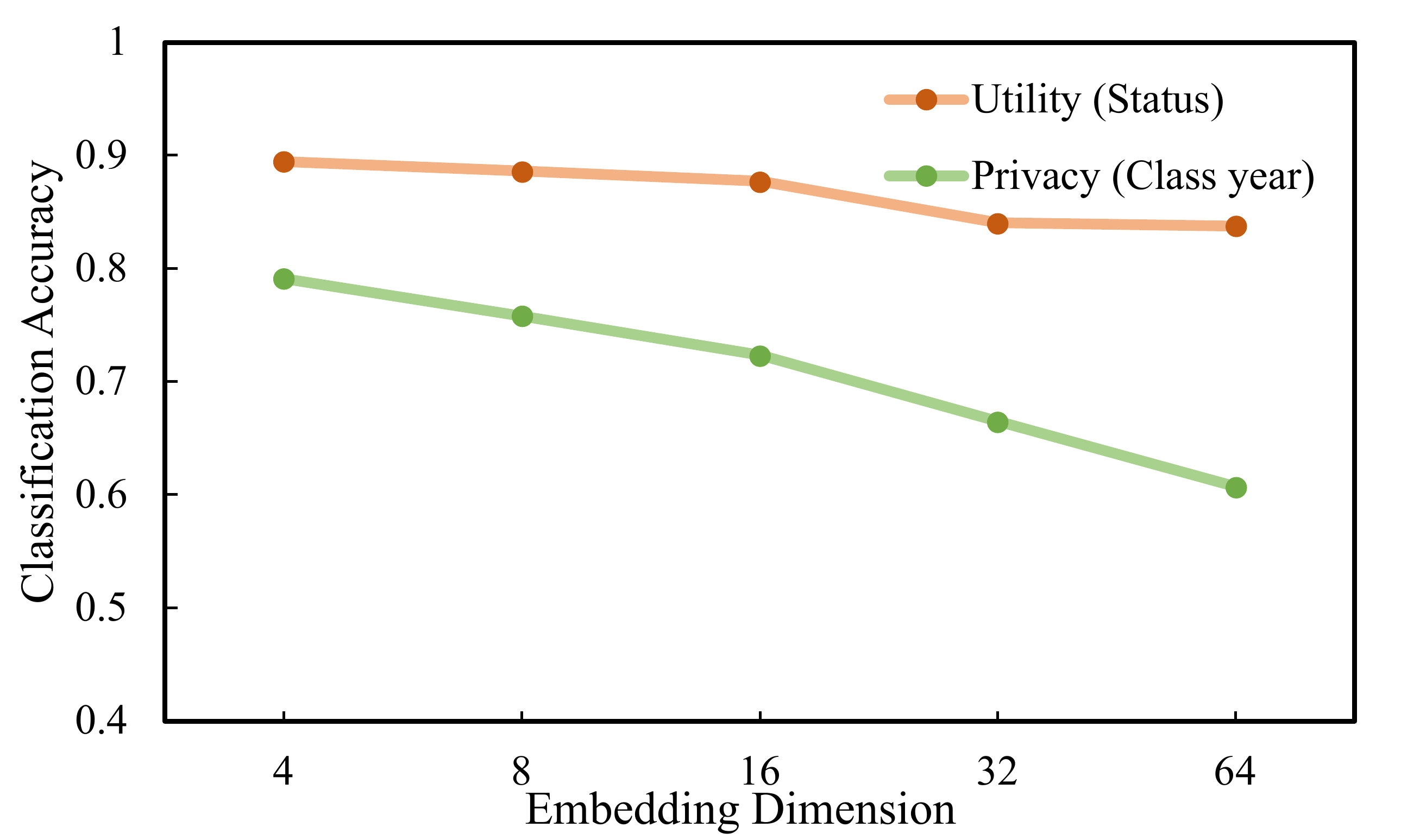}
\label{subfig:dim-yale}}\hfil
\subfigure[Rochester]{\includegraphics[width=0.33\textwidth]{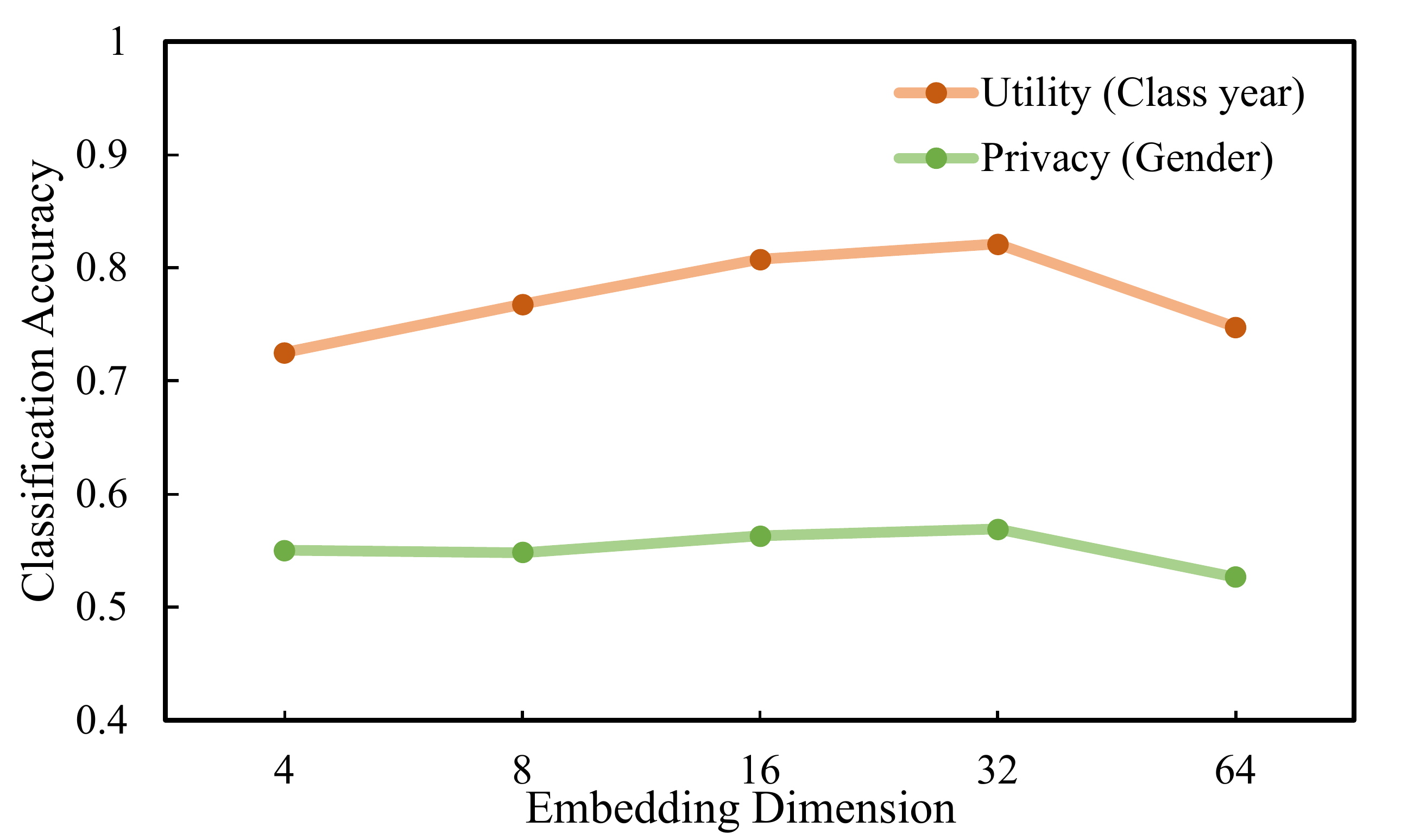}
\label{subfig:dim-rochester}}\hfil 
\subfigure[Credit defaulter]{\includegraphics[width=0.33\textwidth]{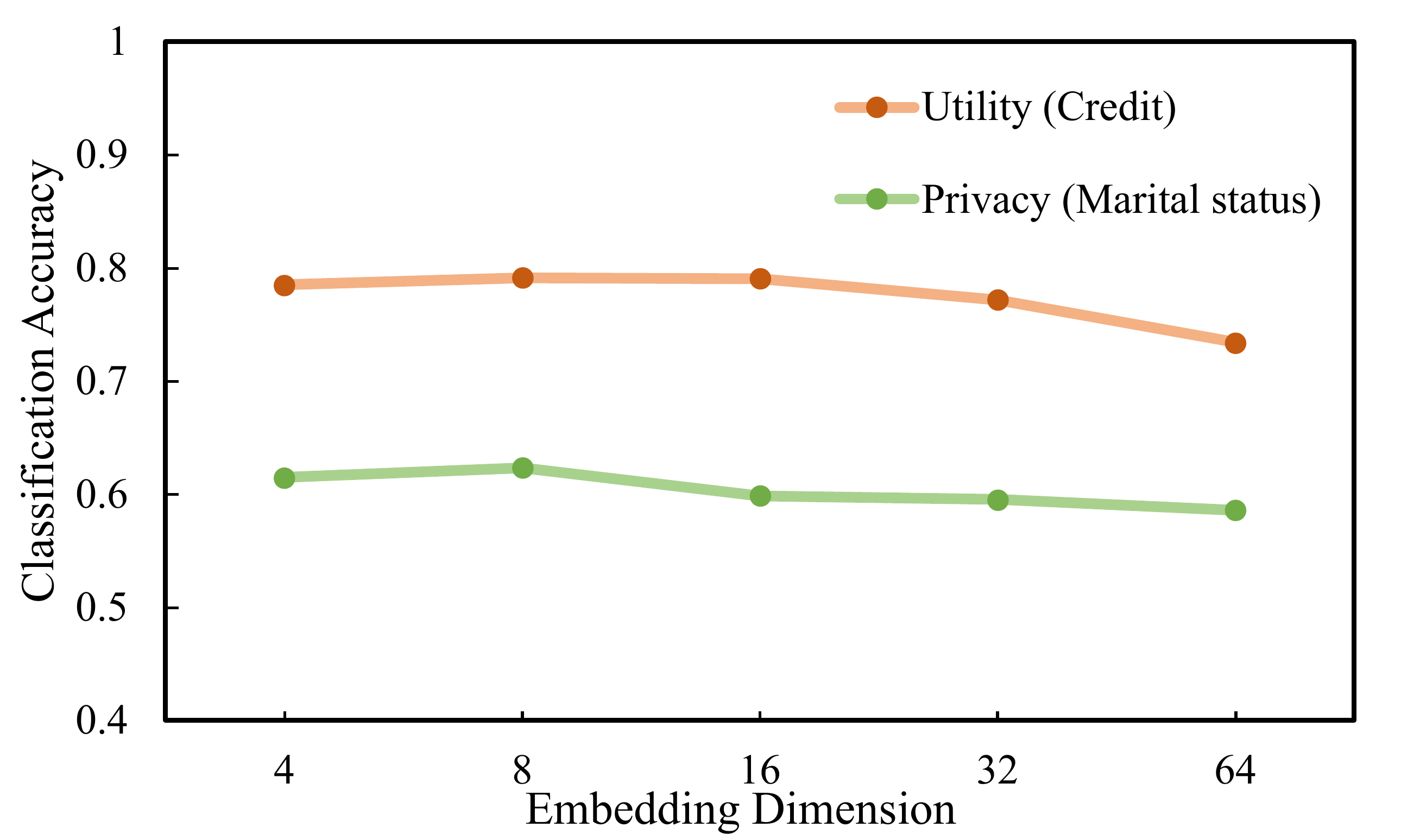}} 
\caption{Utility performance and privacy protection evaluation with different graph embedding dimensions.}
\label{fig:dim}
\end{figure*}

\subsubsection{Partially observed sensitive attributes} However, protecting privacy with fully observed sensitive attributes is not always the case for various reasons. In practice, we can only generate private graph embeddings with partially observed sensitive attributes. Because APGE, GAE-MI, and VFAE have to know sensitive attributes in advance, we only compare PVGAE with DP-GCN under different observed sensitive attribute amounts. We select the observed attribute ratio from $\{0.1, 0.3, 0.5, 0.7, 0.9\}$ and compare the private attribute inference accuracy. The results are shown in Figure \ref{fig:ratio}. Because various observed attributes ratios have minor impacts on the two models' utility performance, which is similar to fully observed sensitive attributes evaluation (Table \ref{tab:main}), we only report the privacy inference accuracy. From the figure, we can know that our proposed PVGAE performs better in privacy protection compared to DP-GCN, as the attacker always has lower prediction accuracy facing embeddings generated by PVGAE. This is because two orthogonal representations do not ensure they are independent. Besides, PVGAE can better resist the influence of the observed sensitive attributes ratio. For example, for Yale shown in Figure \ref{subfig:ratio-yale}, the privacy preserving performance slightly degrades with fewer observed sensitive attributes for PVGAE, while for DP-GCN, the inference accuracy improves to 0.731 when it knows 10\% sensitive attributes in training. The possible reason is that GCN performs well on semi-supervised node classification problems \cite{kipf2016semi}, the privacy encoders can learn well even with a small part of private attributes while DP-GCN cannot disentangle the representation completely with orthogonal constraints when observing a few sensitive attributes.

\subsection{Sensitivity Study \label{subsec:ablation}}
To better understand the property of PVGAE, we conduct ablation studies to learn the impacts of the independence penalty coefficient, embedding dimensions, and attacker models.

\subsubsection{Independence penalty}
In various types of graphs and tasks, privacy protection may have different importance. In this part, we evaluate the impact of the independence penalty coefficient $\beta$. A larger $\beta$ can provide stronger private attribute protection. We select $\beta$ from $\{0.1, 1, 5, 10, 50, 100\}$, and evaluate the node classification utility task and privacy preserving performance, respectively. The results are shown in Figure \ref{fig:penalty}. We can see that the independence distribution penalty can effectively control privacy protection. With a larger $\beta$, PVGAE can provide more robust privacy protection. In Rochester datasets, as shown in Figure \ref{subfig:penalty-rochester}, the private attributes inference accuracy drops from 0.696 to 0.569 when $\beta$ increases from 0.1 to 100. Besides, though stronger privacy protection leads to a drop in utility performance as there are tradeoffs between utility and privacy, PVGAE can provide strong privacy protection with a relatively slight loss of utility performance. Take Yale datasets shown in Figure \ref{subfig:penalty-yale} as an example, when the utility drops from 0.892 to 0.840 as $\beta$ improves, 5.8\% performance loss brings 19.0\% private attributes inference accuracy drops, which are from 0.821 to 0.665. 
\begin{figure*}
\centering
\subfigure[Yale]{\includegraphics[width=0.325\textwidth]{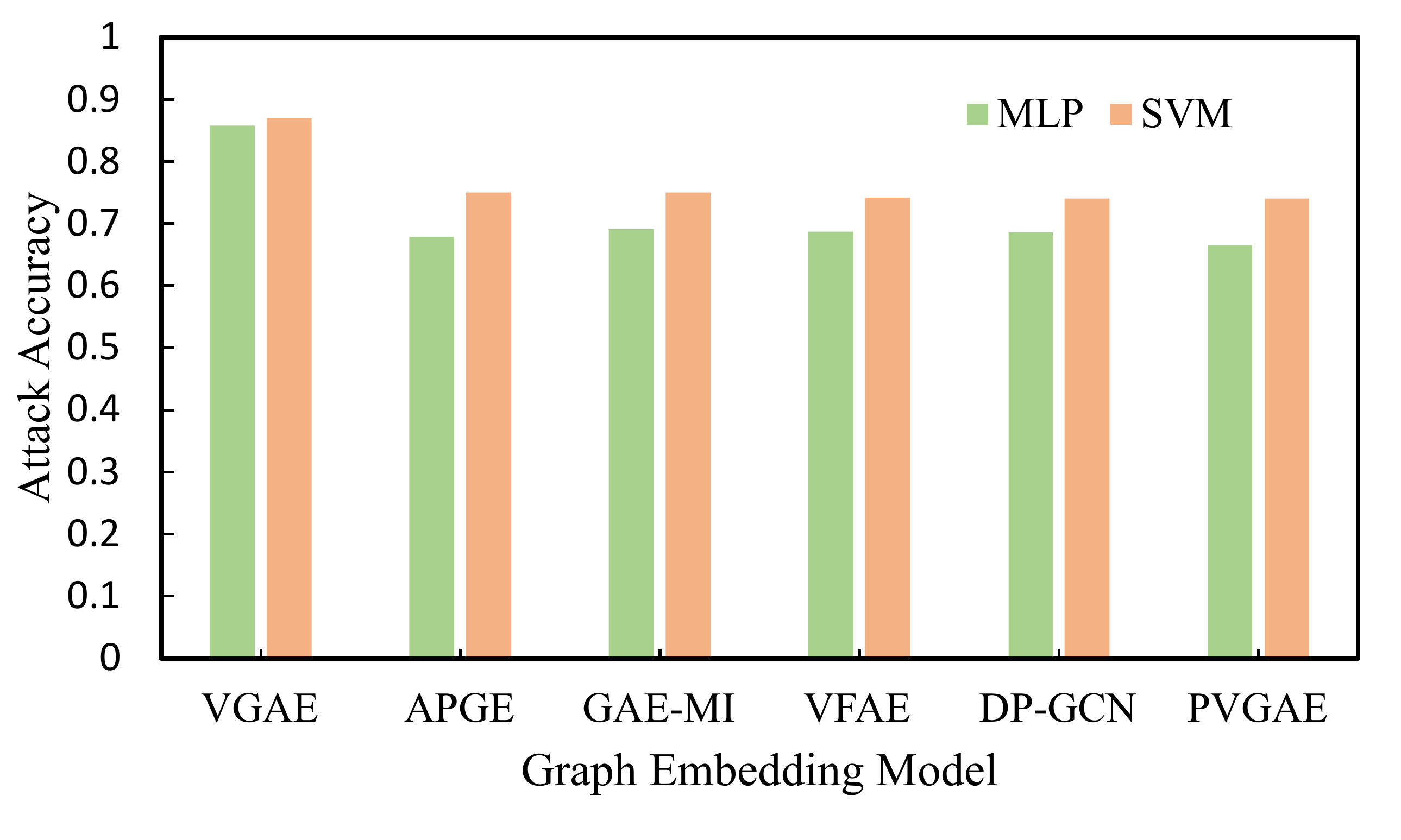}
\label{subfig:attack-yale}}\hfil
\subfigure[Rochester]{\includegraphics[width=0.325\textwidth]{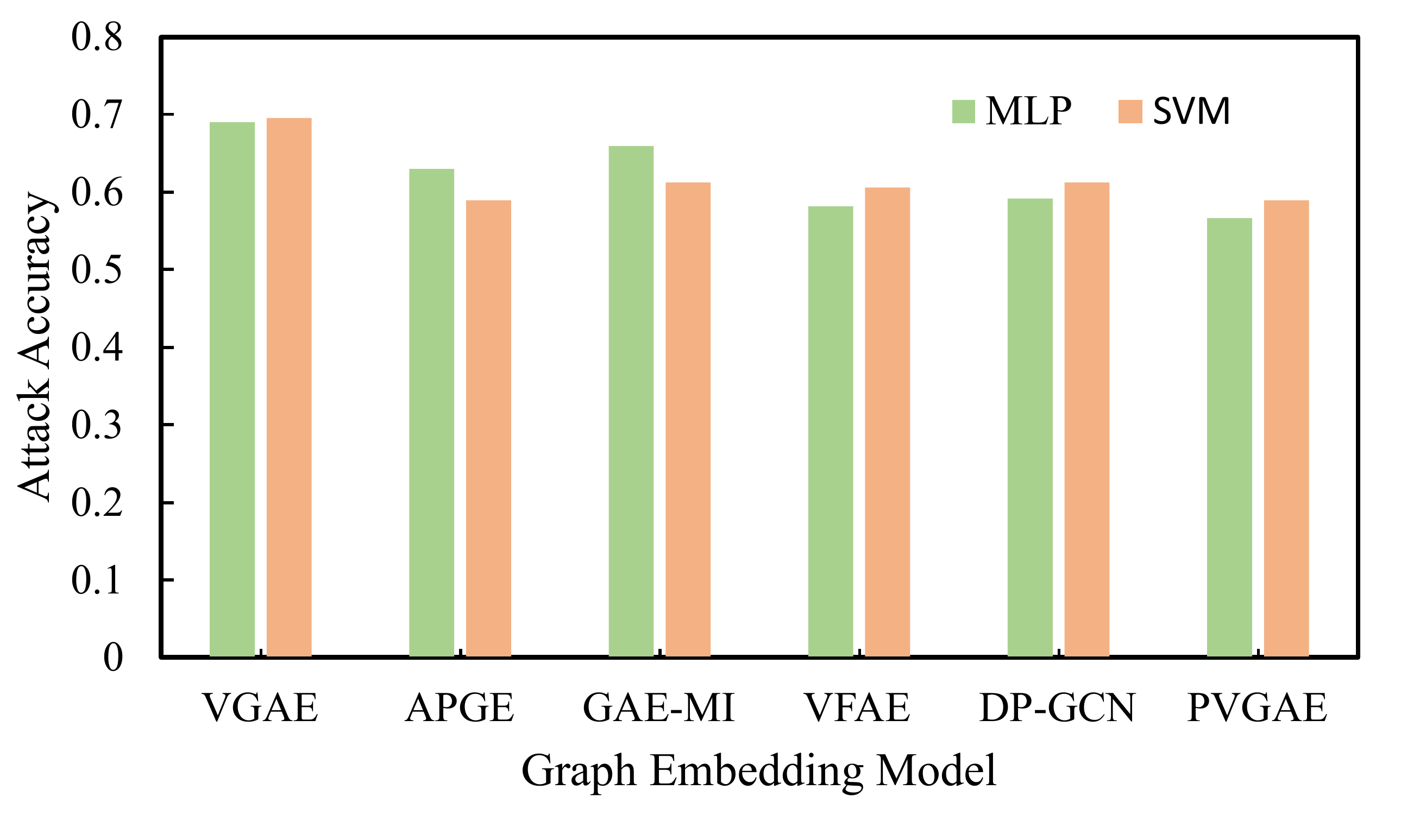}
\label{subfig:attack-rochester}}\hfil 
\subfigure[Credit defaulter]{\includegraphics[width=0.325\textwidth]{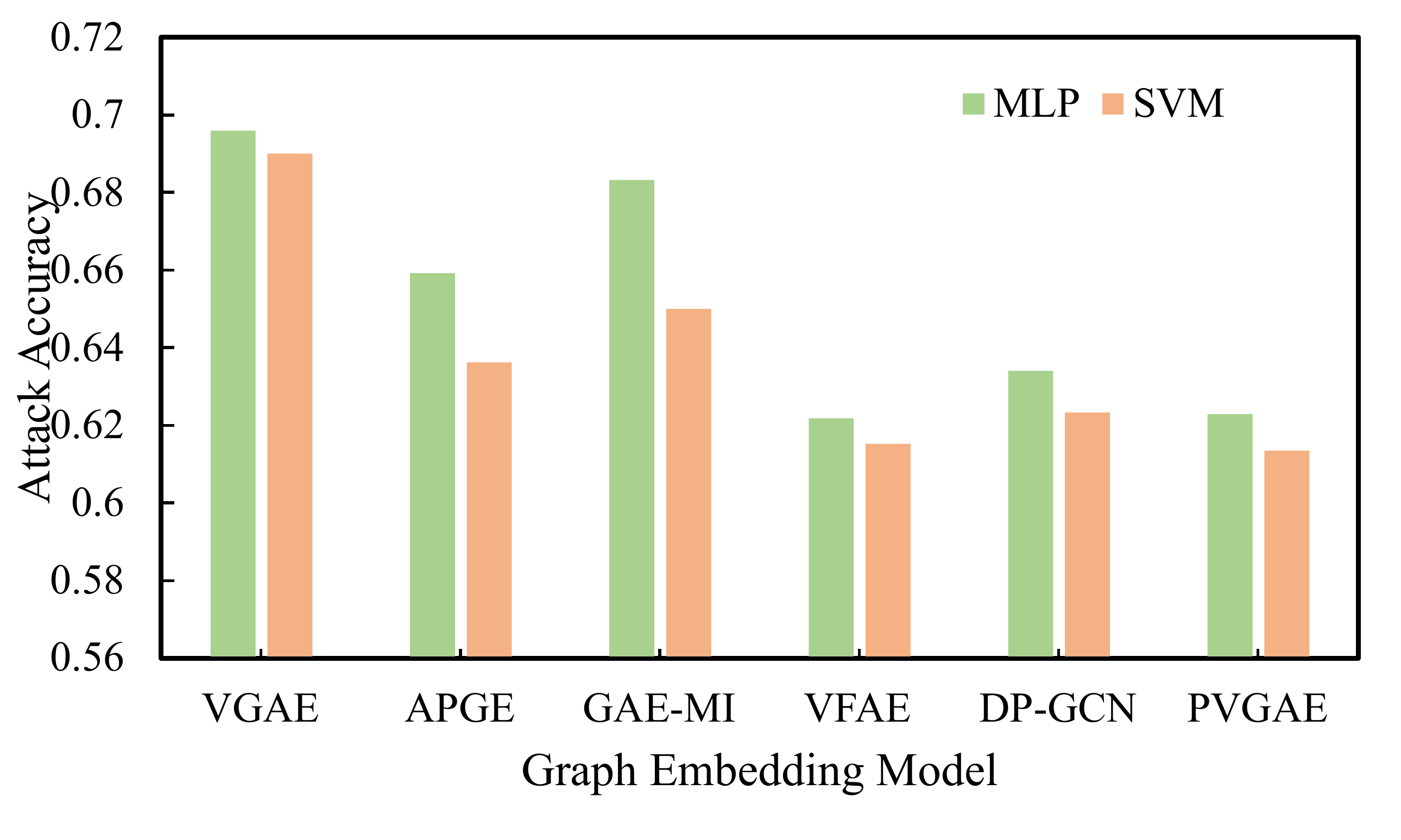}
\label{subfig:attack-credit}} 
\caption{Private attributes inference accuracy with different attack models.}
\label{fig:attack}
\end{figure*} 
\subsubsection{Embedding dimension}
Embedding dimensions largely influence the representation performance and the learned private information. In this part, we evaluate PVGAE's utility and privacy performance with different embedding dimensions. We select embedding dimensions from $\{4, 8, 16, 32, 64\}$ and compare the node classification utility and privacy inference accuracy. From Figure \ref{fig:dim}, we can see that graph embedding dimensions significantly influence the utility and privacy performance. Take the Rochester dataset shown in Figure \ref{subfig:dim-rochester} as an example, the embedding performs best when the graph embedding dimension is 32. Lower and higher dimensions both decrease the representation ability. This may be because lower dimension embeddings' representation ability is limited, and embedding cannot learn the intrinsic relations, which influence both utility performance and private information. While for higher dimension embeddings, they learn some noise in the original graph and harm the downstream tasks.

\subsubsection{Attacker models} Different attack methods may have inconsistent attack results and attacker types may influence the inference results. To validate the attacker model's effectiveness, we apply another classification model SVM to the embeddings generated by all the baselines and PVGAE, and compare the accuracy of the attribute inference. From Figure \ref{fig:attack}, we can see that the attack accuracy of the two classification models is highly correlated, the most significant difference between the two attack models happens in GAE-MI's Credit defaulter graph embeddings as shown in Figure \ref{subfig:attack-credit} where inference accuracy is 0.683 and 0.650 for MLP and SVM respectively, indicating that the embeddings generated by all representation learning methods have similar attack difficulty and MLP's attack accuracy can reflect the privacy leakage degrees. Besides, the privacy preserving performance of our PVGAE is comparable to other baselines for both MLP and SVM attackers. Take Figures \ref{subfig:attack-yale}, \ref{subfig:attack-credit} as an example, we can observe that there exist differences for MLP and SVM prediction accuracy, SVM performs better on Yale while MLP performs better on Credit defaulter. However, in the same datasets, attack difficulty is consistent for MLP and SVM. Therefore, MLP's attack accuracy can denote attack difficulty.  
  
\subsubsection{Impact of different privacy preferences.} We can access sensitive attributes partially due to different privacy preferences and utilize those shared attributes to remove private information. However, utilizing sensitive information itself may cause private information leakage. To evaluate the impact of sharing sensitive attributes, we compare the utility-privacy difference between nodes that share their sensitive attributes (Public) and nodes that keep their sensitive information secret (Secret). We assume that we have observed 50\% sensitive attributes in the training stage and compare those "Public" nodes and "Secret" nodes with utility performance and privacy protection. We assume that an attacker can randomly access part of the sensitive attributes for inference model training. In Table \ref{tab:ob_ratio}, we can observe that "Public" and "Secret" have similar node classification accuracy which indicates similar utility performance, while for privacy protection, an attacker can get higher inference accuracy on "Secret" nodes than "Public" nodes, which indicates that sharing sensitive attributes for model training can enjoy better privacy protection. It may be because the privacy encoder can predict "Public" nodes' sensitive attributes more accurately as it observes these attributes directly, therefore, we can better disentangle the private information.
\begin{table}[t]
\caption{Utility and privacy evaluation for difference privacy preferences.}
\label{tab:ob_ratio}
\begin{tabular}{c|cc|cc}
\hline
\multirow{2}{*}{Dataset} & \multicolumn{2}{c|}{Node Classification (ACC)} & \multicolumn{2}{c}{Privacy (ACC)} \\ \cline{2-5} 
                         & Public                & Secret                & Public          & Secret         \\ \hline
Yale                     & 0.838                  & 0.837                 & 0.676            & 0.688          \\
Rochester                & 0.808                  & 0.812                 & 0.570             & 0.571          \\
Credit                   & 0.782                  & 0.779                 & 0.627            & 0.635          \\ \hline
\end{tabular}

\end{table}


\section{Conclusion \label{sec:conclude}}
In this paper, we present a novel distribution regularization for graph autoencoders to generate private graph embeddings against attribute inference attacks. PVGAE utilizes independence regularization to disentangle the utility representation distribution with sensitive information. Besides, our framework can be applied with partially observed sensitive attributes so that PVGAE has a greater scope of application compared to most privacy preserving representation learning methods. We theoretically discuss the effectiveness of the proposed regularization under the mutual information perspective. Experimental results on three real-world datasets demonstrate that the proposed model has competitive privacy preserving ability and utility performance compared to other privacy preserving representation learning with fully observed sensitive attributes. While in partially observed attributes scenarios, our model can provide a better utility-privacy tradeoff compared to existing works. In the future, as we only consider one sensitive attribute in PVGAE, we will extend our regularization to protect multiple private attributes.

\section*{ACKNOWLEDGMENTS}
The authors of this paper were supported by the NSFC Fund (U20B2053) from the NSFC of China, the RIF (R6020-19 and R6021-20) and the GRF (16211520 and 16205322) from RGC of Hong Kong, the MHKJFS (MHP/001/19) from ITC of Hong Kong and the National Key R\&D Program of China (2019YFE0198200) with special thanks to HKMAAC and CUSBLT. We also thank the UGC Research Matching Grants (RMGS20EG01-D, RMGS20CR11, RMGS20CR12, RMGS20EG19, RMGS20EG21, RMGS23CR05, RMGS23EG08).

\bibliographystyle{ACM-Reference-Format}
\bibliography{sample-base}

\end{document}